\documentclass[10pt,twocolumn,letterpaper]{article}

\usepackage{cvpr}
\usepackage{times}
\usepackage{epsfig}
\usepackage{graphicx}
\usepackage{amsmath}
\usepackage{amssymb}

\usepackage{algorithm}
\usepackage{algorithmic}
\usepackage{stfloats}
\usepackage{amsfonts}
\usepackage{slashbox}
\usepackage{pdfpages}
\usepackage{multirow}
\usepackage{subfigure}

\usepackage[breaklinks=true,bookmarks=false]{hyperref}

\cvprfinalcopy 

\def\cvprPaperID{****} 

\ifcvprfinal\fi
\begin{document}

\title{Video Captioning with Transferred Semantic Attributes}


\author{Yingwei Pan $^{\dag}$, Ting Yao $^{\ddag}$, Houqiang Li $^{\dag}$, and Tao Mei $^{\ddag}$ \\
{\small\centering$^{\dag}$University of Science and Technology of China, Hefei, China}~~~~
{\small\centering$^{\ddag}$Microsoft Research, Beijing, China}\\
{\tt\small panyw.ustc@gmail.com, \{tiyao, tmei\}@microsoft.com, lihq@ustc.edu.cn}
}

\maketitle
\thispagestyle{empty}

\begin{abstract}
Automatically generating natural language descriptions of videos plays a fundamental challenge for computer vision community. Most recent progress in this problem has been achieved through employing 2-D and/or 3-D Convolutional Neural Networks (CNN) to encode video content and Recurrent Neural Networks (RNN) to decode a sentence. In this paper, we present Long Short-Term Memory with Transferred Semantic Attributes (LSTM-TSA)---a novel deep architecture that incorporates the transferred semantic attributes learnt from images and videos into the CNN plus RNN framework, by training them in an end-to-end manner. The design of LSTM-TSA is highly inspired by the facts that 1) semantic attributes play a significant contribution to captioning, and 2) images and videos carry complementary semantics and thus can reinforce each other for captioning. To boost video captioning, we propose a novel transfer unit to model the mutually correlated attributes learnt from images and videos. Extensive experiments are conducted on three public datasets, i.e., MSVD, M-VAD and MPII-MD. Our proposed LSTM-TSA achieves to-date the best published performance in sentence generation on MSVD: 52.8\% and 74.0\% in terms of BLEU@4 and CIDEr-D. Superior results when compared to state-of-the-art methods are also reported on M-VAD and MPII-MD.
\end{abstract}

\section{Introduction}
Video captioning, which is known as describing videos with natural language, has brought a profound challenge to both computer vision and language processing communities. Intensive research interests have been paid for this emerging topic. Existing approaches to video captioning have evolved through two dimensions: template-based language model \cite{Guadarrama:ICCV13,Rohrbach:ICCV13,Xu:AAAI15} and sequence learning method \cite{Pan:CVPR16,Venugopalan:ICCV15,Yao:ICCV15,Yu:CVPR16}. The former predefines a set of templates for sentence generation following specific grammar rules and aligns each part of sentence with image content. This category of model, however, highly depends on the pre-defined templates and thus the generated sentences are always with constant syntactical structure. Sequence learning method, in contrast, is to leverage sequence learning models to directly translate video content into sentence, which is mainly inspired from the recent advances by using RNNs in machine translation \cite{Sutskever:NIPS14}. The spirit behind is an encoder-decoder mechanism for translation. More specifically, an encoder 2-D/3-D Convolutional Neural Networks (CNN) reads a video and produces a vector of video representations, which in turn is fed into a decoder Recurrent Neural Networks (RNN) that generate a natural sentence. While encouraging performances are reported, the CNNs plus RNNs based sequence learning approaches translate directly from video representations to language, leaving the high-level semantic cues in the video under explored. However, the utilization of high-level semantic information, i.e., semantic attributes, has shown effective in the vision to language tasks \cite{Wu:CVPR16} (e.g., image captioning and visual Q\&A).

\begin{figure}[!tb]
\centering {\includegraphics[width=0.48\textwidth]{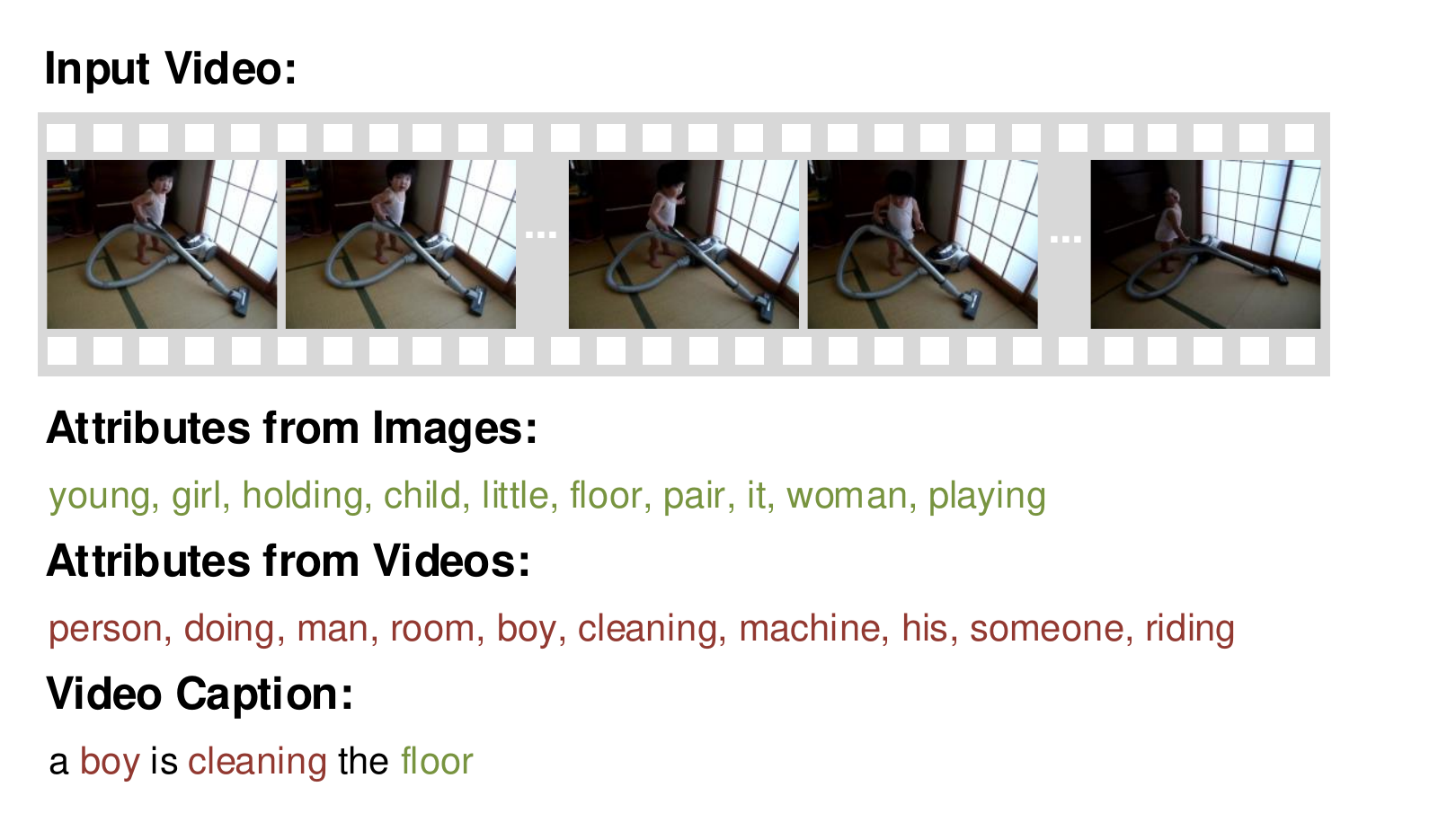}}
\caption{\small An example of video description generation. The input is a short video clip and the attributes are learnt from images and videos, respectively. The output is a sentence generated by our LSTM-TSA architecture.}
\label{fig:figAttr}
\vspace{-0.10in}
\end{figure}

This paper proposes a novel deep architecture, named Long Short-Term Memory with Transferred Semantic Attributes (LSTM-TSA), which takes advantages of incorporating semantic attributes into sequence learning for video captioning. More importantly, take the given video in Figure \ref{fig:figAttr} as an example, the semantic properties observed in images often depict static objects and scenes (e.g., ``girl," ``child," and ``floor") while the semantic cues extracted from videos convey the temporal dynamics more (e.g., ``doing," ``cleaning" and ``riding"). This has made the attributes mined from images and videos complementary to generate the sentence for the video (e.g., ``a boy is cleaning the floor"). We investigate particularly how the attributes from two sources are fused and leveraged for enhancing video captioning. Specifically, given a video, a 2-D/3-D CNN is utilized to extract visual features of selected video frames/clips and the video representations are produced by mean pooling over these visual features. Then, a LSTM for generating video description is learnt by feeding into both video representations and semantic attributes mined from images and videos. To better leverage the attributes from two sources, a transfer unit is devised to dynamically balance the influence in between given the input word and the hidden state in LSTM.

The main contribution of this work is the proposal of LSTM-TSA for addressing the issue of exploiting the mutual relationship between video representations and attributes for boosting video captioning. This issue also leads to an elegant view of how complementary attributes from images and videos are jointly exploited for sentence generation, which is a problem not yet fully explored in the literature.

\section{Related Work}\label{sec:RW}
We briefly group the related works into two categories: video captioning and sequence learning by using attributes. The former draws upon research in automatically generating description to a video, and the later investigates sequence learning for visual content by utilizing the attributes.

\textbf{Video Captioning.} The research in this direction has proceeded along two different dimensions: template-based language methods \cite{Guadarrama:ICCV13,Kojima:IJCV02,Rohrbach:ICCV13,Xu:AAAI15} and sequence learning approaches (e.g., RNNs) \cite{Pan:CVPR16,Venugopalan:ICCV15,Venugopalan:NAACL15,Yao:ICCV15,Yu:CVPR16}. Template-based language methods firstly align each sentence fragments (e.g., subject, verb, object) with detected words from visual content and then generate the sentence with predefined language template. Obviously, most of them highly depend on the templates of sentence and always generate sentence with syntactical structure. \cite{Kojima:IJCV02} is one of the earlier works that builds a concept hierarchy of actions for natural language description of human activities. Rohrbach \emph{et al.} learn a CRF to model the relationships between different components of the input video and generate description for video \cite{Rohrbach:ICCV13}. Recently, a deep joint video-language embedding model in \cite{Xu:AAAI15} is designed for video sentence generation. Different from template-based language methods, sequence learning approaches learn the probability distribution in the common space of visual content and textual sentence to generate novel sentences with more flexible syntactical structure. In \cite{Venugopalan:NAACL15}, Venugopalan \emph{et al.} present a LSTM based model to generate video descriptions with the mean pooling representation over all frames. The framework is then extended by inputting both frames and optical flow images into an encoder-decoder LSTM in \cite{Venugopalan:ICCV15}. Furthermore, Pan \emph{et al.} additionally consider the relevance between sentence semantics and video content as a regularizer in LSTM based architecture \cite{Pan:CVPR16}. Compared to mean pooling, Yao \emph{et al.} propose to utilize the temporal attention mechanism to exploit temporal structure for video captioning \cite{Yao:ICCV15}.

\textbf{Sequence Learning by Using Attributes.} Attributes are properties observed in visual content with rich semantic cues and have been widely studied in computer vision for improving the efficacy of visual recognition \cite{Devi:ICCV11}. Following this elegant recipe, several recent works have attempted to inject attributes into sequence learning for image caption generation. Fang \emph{at al.} \cite{Fang:CVPR15} firstly use Multiple Instance Learning to train attributes detector and then generate sentence through a maximum-entropy language model based on the outputs of attributes detector. Later in \cite{Tran2016rich}, this framework is further developed with a larger range of attributes, additionally including celebrities and landmarks, to enrich the generated sentence. More recently, in \cite{Wu:CVPR16}, high-level concepts/attributes are shown to obtain clear improvements on image captioning task when injected into existing state-of-the-art RNN-based model and such visual attributes are also utilized as semantic attention in \cite{You:CVPR16} to enhance image captioning.

In summary, our work presents the first effort to leverage semantic attributes in video captioning. Different from most of the aforementioned sequence learning models using attributes which mainly focus on sentence generation by solely depending on the attributes learnt in domain, our work contributes by studying not only learning attributes in videos from both image and video domains, but also how the attributes can be better fused by dynamically offering a transfer unit in between for boosting video captioning.

\section{Video Captioning with Transferred Semantic Attributes}\label{sec:VCSA}
In this paper, we devise our CNN plus RNN architecture to generate video descriptions under the umbrella of incorporating mined semantic attributes from images and videos. Specifically, we begin this section by presenting the problem formulation and how to learn semantic attributes in videos, followed by our proposed LSTM-TSA video captioning framework. In particular, several variants of our designed transfer unit which is utilized to fuse the attributes learnt from two sources are investigated and discussed.

\subsection{Problem Formulation}
Suppose we have a video $V$ with $N_v$ sample frames/clips (uniform sampling) to be described by a textual sentence $\mathcal {S}$, where $\mathcal{S} = \{w_1, w_2, ..., w_{N_s}\}$ consisting of $N_s$ words. Let ${\bf{v}}\in {\mathbb{R}}^{D_v}$ and ${\bf{w}}_t\in {{\mathbb{R}}^{D_w}}$ denote the $D_v$-dimensional video representations of the video $V$ and the $D_w$-dimensional textual features of the $t$-th word in sentence $\mathcal{S}$, respectively. As a sentence consists of a sequence of words, a sentence can be represented by a $D_w\times N_s$ matrix ${\bf{W}}\equiv [{\bf{w}}_1, {\bf{w}}_2,...,{\bf{w}}_{N_s}]$, with each word in the sentence as its column vector. Furthermore, we have another two feature vectors ${\bf{A_i}}\in {\mathbb{R}}^{D_{a_i}}$ and ${\bf{A_v}}\in {\mathbb{R}}^{D_{a_v}}$ to represent the probability distribution over the high-level attributes for video $\mathcal {V}$ learnt from images and videos, respectively. More details about how we mine and represent the attributes from images and videos will be introduced in Section \ref{Sec:SAIV}.

Inspired by the recent successes of probabilistic sequence models leveraged in statistical machine translation \cite{Bahdanau14, Sutskever:NIPS14} and semantic attributes utilized in image captioning \cite{Fang:CVPR15,You:CVPR16}, we aim to formulate our video captioning model in an end-to-end fashion based on LSTM \cite{Hochreiter:NC97} which encodes the given video and its learnt attributes from both images and videos into a fixed dimensional vector and then decodes it to the output target sentence. Hence, the video sentence generation problem we exploit here can be formulated by minimizing the following energy loss function as
\begin{equation}\label{Eq:EqPF1}\small
E({\bf{v}}, {\bf{A_i}}, {\bf{A_v}}, {\mathcal {S}}) = -\log {\Pr{({\mathcal {S}}|{\bf{v}}, {\bf{A_i}}, {\bf{A_v}})}},
\end{equation}
which is the negative $\log$ probability of the correct textual sentence given the video and detected attributes from both images and videos.

Since the model produces one word in the sentence at each time step, it is natural to apply chain rule to model the joint probability over the sequential words. Thus, the $\log$ probability of the sentence is given by the sum of the $\log$ probabilities over the word and can be expressed as
\begin{equation}\label{Eq:EqPF2}\scriptsize
\log {\Pr{({\mathcal {S}}|{\bf{v}}, {\bf{A_i}}, {\bf{A_v}})}} =  \sum\limits_{t = 1}^{{N_s}} {\log \Pr\left( {\left. {{{\bf{w}}_t}} \right|{\bf{v}}, {\bf{A_i}}, {\bf{A_v}}, {{\bf{w}}_0}, \ldots ,{{\bf{w}}_{t - 1}}} \right)}.
\end{equation}
By minimizing this loss, the contextual relationship among the words in the sentence can be guaranteed given the video and its learnt attributes from images and videos.

\begin{figure}[!tb]
\centering {\includegraphics[width=0.5\textwidth]{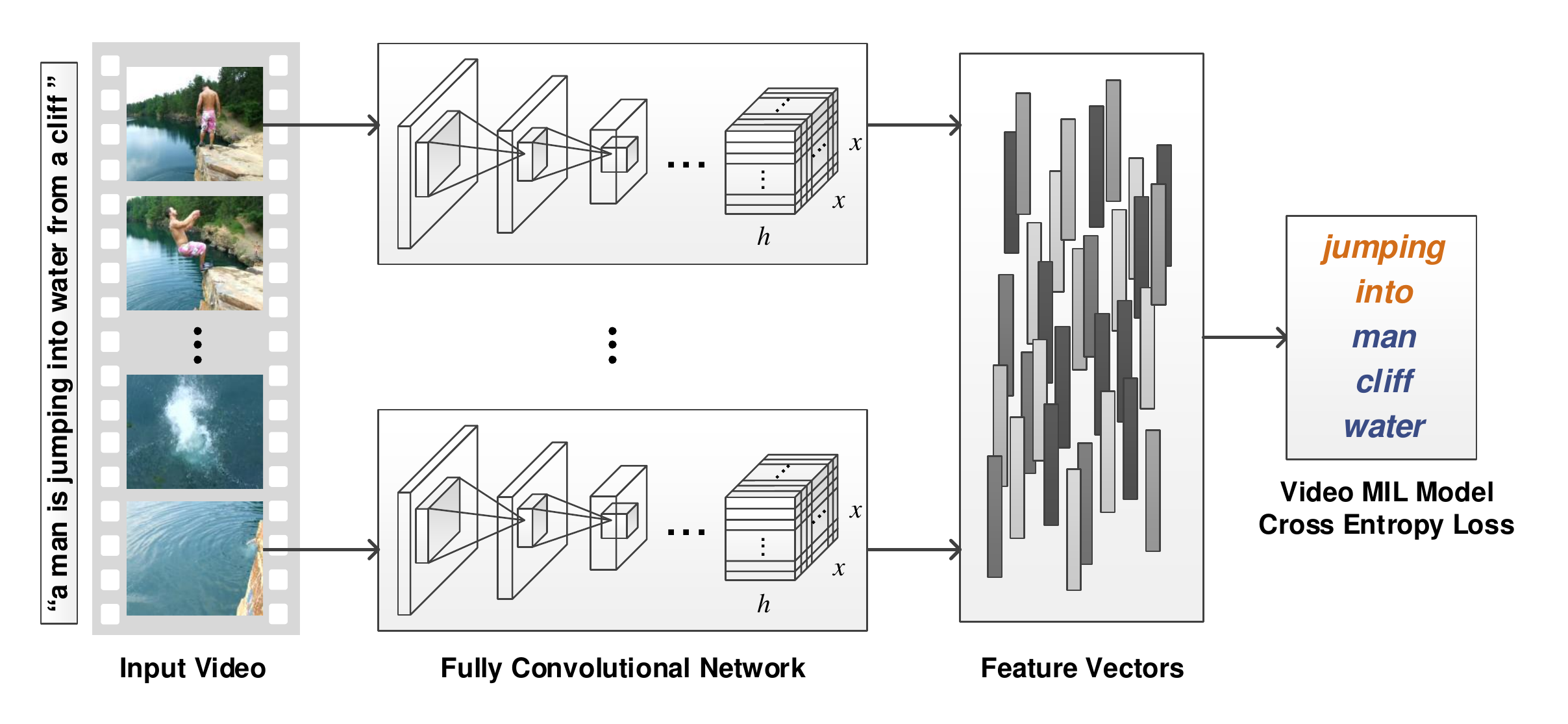}}
\caption{Video MIL framework.}
\label{fig:figMIL}
\vspace{-0.1in}
\end{figure}

\subsection{Semantic Attributes in Video}\label{Sec:SAIV}
\textbf{Attributes Learnt from images.} We draw inspiration from recent advances in attribute detection for image captioning \cite{Fang:CVPR15,You:CVPR16} and adopt the weakly-supervised approach of Multiple Instance Learning (MIL) on image captioning benchmarks (e.g., COCO \cite{Lin:ECCV14}) to learn attribute detectors. For an attribute $w_a$, one image $I$ is regarded as a positive bag of regions (instances) if $w_a$ exists in image $I$'s ground-truth sentences, and negative bag otherwise. By inputting all the bags into a noisy-OR MIL model \cite{Zhang:NIPS05}, the probability of the bag $b_I$ which contains attribute $w_a$ is measured on the probabilities of all the regions in the bag as
\begin{equation}\label{Eq:SAIV1}\small
{\Pr}_{I}^{{w_a}} = 1 - \prod\limits_{{r_i} \in {b_I}} {\left( {1 - p_i^{{w_a}}} \right)} ,
\end{equation}
where $p_i^{{w_a}}$ is the probability of the attribute $w_a$ predicted by region $r_i$ and can be calculated through a sigmoid layer after the last convolutional layer in the CNN architecture \cite{Fang:CVPR15} which is a fully convolutional network extended from recent popular CNN \cite{Simonyan14}. Specifically, the dimension of convolutional activations from the last convolutional layer is $x \times x \times h$ and $h$ represents the representation dimension of each region, resulting in $x \times x$ response map which preserves the spatial dependency of the image. Then, a cross entropy loss is calculated based on the probabilities of all the attributes at the top of the whole architecture to optimize image MIL model. With the learnt image MIL model on image captioning dataset, we compute the probability distribution on all the attributes for each sampled frame and perform mean pooling over distributions of all the sampled frames to obtain the final representations ${\bf{A_i}}$ of attributes learnt from images.

\textbf{Attributes Learnt from videos.} To detect attributes from videos, one natural way is to directly train image MIL model on video frames. However, as a video is a sequence of frames with large variations, simply assigning video-level description to each sampled frame will lead to the issue of semantics shift and thus involve noise in the process of attribute learning. To solve this problem, a video MIL model is particularly devised to learn attributes from videos, as shown in Figure \ref{fig:figMIL}.

Given an attribute $w_a$, we treat the spatial regions of all the $N_V$ sampled frames in video $V$ as one bag, which is considered as positive if $w_a$ exists in video $V$'s descriptions and negative otherwise. By feeding all the bags into the fully convolutional network with the same architecture in image MIL model, we calculate the probability of bag $b_V$ which contains attribute $w_a$ on the probabilities of all the regions in the bag as
\begin{equation}\label{Eq:SAIV2}\small
{\Pr}_V^{{w_a}} = 1 - \prod\limits_{j \in \left[ {1,{N_V}} \right]} {\prod\limits_{{r_{ij}} \in b_V^{\left( j \right)}} {\left( {1 - p_{ij}^{{w_a}}} \right)} } ,
\end{equation}
where $p_{ij}^{{w_a}}$ is the probability of the attribute $w_a$ predicted by the $i$-th region in the $j$-th frame and $b_V^{\left( j \right)}$ denotes the set of all the regions in the $j$-th frame. Specifically, in our training, all the $N_V$ sampled frames from one video are taken as a batch and each frame is fed into the same fully convolutional network followed by a sigmoid layer, resulting in $x \times x$ response map whose element represents the probability $p_{ij}^{{w_a}}$ of attribute $w_a$ detected in region $r_{ij}$. Similar to image MIL model, a cross entropy loss layer is designed at the top of the whole architecture to optimize our video MIL model. As such, the proposed video MIL model is trained holistically among all the frames in the video and the probability distribution calculated by Eq.(\ref{Eq:SAIV2}) are employed as representations ${\bf{A_v}}$ of attributes learnt from videos.

\subsection{Video Captioning with Semantic Attributes Learnt from Images and Videos}
With the detected high-level semantic attributes learnt from images and videos, we propose a Long Short-Term Memory with Transferred Semantic Attributes from Images and Videos (LSTM-TSA$_{IV}$) model for video captioning. The basic idea of LSTM-TSA$_{IV}$ is to translate the video representation from a 2-D and 3-D CNN to the desired output sentence through LSTM-type RNN model by additionally injecting the high-level semantic attributes learnt from both images and videos. Specifically, a transfer unit is designed to dynamically control the impacts of semantic attributes from the two sources on sentence generation.

Next, we will first present the architecture of our attributes-based LSTM-type captioning model, followed by introducing the designed transfer unit and how to integrate it into LSTM for video captioning.

\subsubsection{Attributes-based LSTM-type Video Captioning}
\begin{figure}[!tb]
\centering {\includegraphics[width=0.5\textwidth]{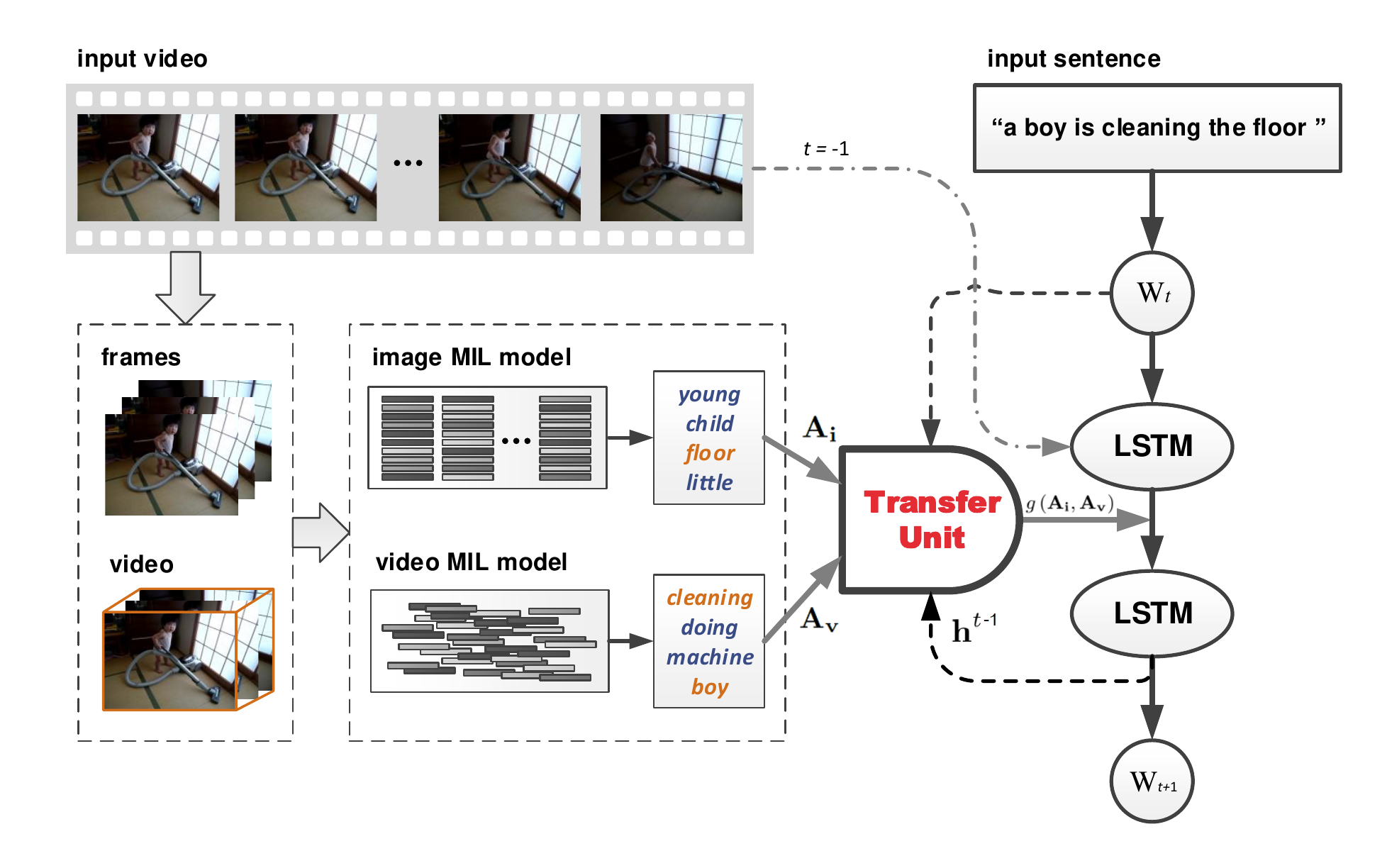}}
\caption{\small The overview of Long Short-Term Memory with Transferred Semantic Attributes (LSTM-TSA) for video captioning (better viewed in color). The video representation is produced by mean pooling over the visual features of sampled frames/clips extracted by a 2-D/3-D CNN, which is injected into LSTM only at the initial time. Image and video MIL models are used to mine semantic attributes from images and videos respectively, which are additionally incorporated into LSTM for boosting video captioning. To better leverage the mined attributes from two sources, a transfer unit is devised to dynamically fuse them into LSTM.}
\label{fig:figSA}
\vspace{-0.1in}
\end{figure}

Inspired by the best-performing architecture (factored, two-layer LSTM) in LRCN \cite{Donahue14}, we devise our attributes-based LSTM-type video captioning model by injecting both video representation and its detected semantic attributes learnt from images and videos into LSTM, as illustrated in Figure \ref{fig:figSA}. In particular, our LSTM-TSA$_{IV}$ model firstly encodes video representation ${\bf{v}}$ at the initial step and then feeds attributes representations from images and videos as the additional inputs to the second-layer LSTM unit at each time step to emphasize the semantic information more frequently. The LSTM updating procedure in LSTM-TSA$_{IV}$ is as
\begin{eqnarray}\label{Eq:Eqlstm}\small
{{\bf{x}}^{-1}} = f_1\left({{\bf{T}}_{v}}{\bf{v}}\right)+g\left({{\bf{A_i}}},{{\bf{A_v}}}\right),\quad\quad\quad\quad\quad\quad\quad\quad\quad~~\\
{{\bf{x}}^t} = f_1\left({{\bf{T}}_s}{{\bf{w}}_t}\right)+g\left({{\bf{A_i}}},{{\bf{A_v}}}\right) ,t \in \left\{ {0, \ldots ,{N_s}-1} \right\},\\
{{\bf{h}}^{t}} = f_2\left( {{{\bf{x}}^t}} \right),t \in \left\{ {0, \ldots ,{N_s}-1} \right\},\qquad\qquad\qquad\quad
\end{eqnarray}
where $D_e$ is the dimension of LSTM input, ${{\bf{T}}_v} \in {{\mathbb{R}}^{{D_e} \times {D_v}}}$ and ${{\bf{T}}_s} \in {{\mathbb{R}}^{{D_e} \times {D_w}}}$ are the transformation matrices for video representation and textual features of word, ${\bf{x}}^t$ and ${\bf{h}}^t$ are the inputs and cell output of the second-layer LSTM unit, $f_1$ and $f_2$ are the updating functions within the first/second-layer LSTM units, and $g$ is the transformation function to transfer both $\bf{A_i}$ and $\bf{A_v}$ into the second-layer LSTM unit.

\subsubsection{Transfer Unit}\label{Sec:TG}
To contextually transfer the information of semantic attributes from multiple sources into LSTM, we devise a novel transfer unit, which is treated as the core unit in our proposed LSTM-TSA$_{IV}$ model.

\begin{figure}[!tb]
\centering {\includegraphics[width=0.49\textwidth]{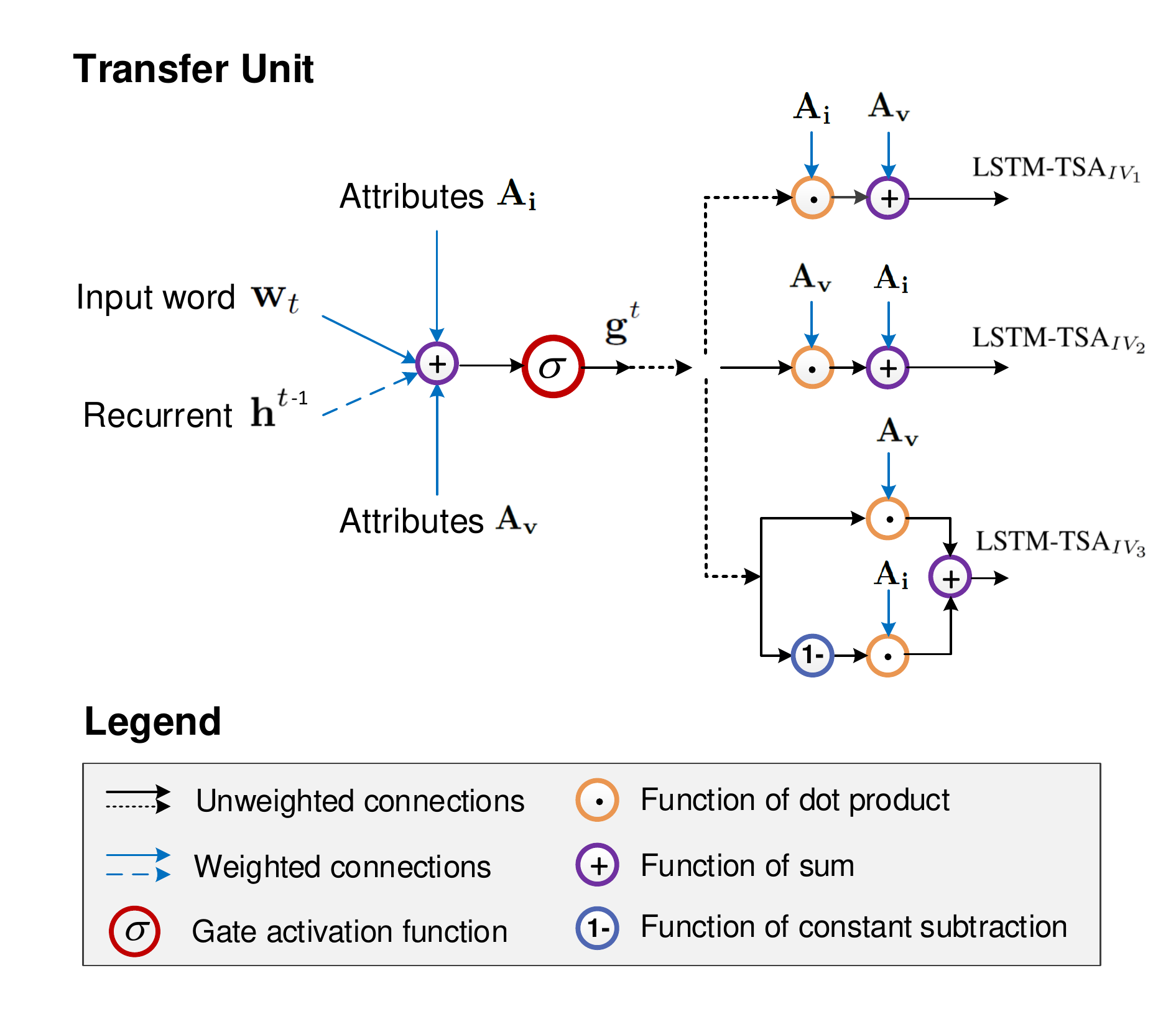}}
\caption{\small Three different architectures of transfer unit with transfer gate (left side) in our LSTM-TSA$_{IV}$ framework.}
\label{fig:figTG}
\vspace{-0.08in}
\end{figure}

\paragraph{Transfer Gate.} A novel gate architecture, named as transfer gate, is especially designed to control the impact of semantic attributes by taking contextual information into account, which is the left part of transfer unit as shown in Figure \ref{fig:figTG}. At the $t$-th time step, the transfer gate encapsulates both the static information (attributes learnt from images and videos) and dynamic (contextual) information (current input word and previous LSTM hidden state) to select valuable knowledge from attributes, which is applied with feature transformation, to produce a fix-length weight vector and followed by a sigmoid function to squash the real-valued weight vector to a range of $[0,1]$. Such output weight vector ${\bf{g}}^t$ for transfer gate is computed as
\begin{equation}\label{Eq:Eqtg1}\small
{{\bf{g}}^t} = \sigma ({{\bf{G}}_s}{{\bf{w}}_t} + {{\bf{G}}_h}{{\bf{h}}^{t - 1}} + {{\bf{G}}_{i}}{\bf{A_i}}+{{\bf{G}}_{v}}{\bf{A_v}}),
\end{equation}
where $D_h$ is the dimension of LSTM cell output, ${{\bf{G}}_s} \in {{\mathbb{R}}^{{D_e} \times {D_{w}}}}$, ${{\bf{G}}_h} \in {{\mathbb{R}}^{{D_e} \times {D_{h}}}}$, ${{\bf{G}}_i} \in {{\mathbb{R}}^{{D_e} \times {D_{a_i}}}}$ and ${{\bf{G}}_v} \in {{\mathbb{R}}^{{D_e} \times {D_{a_v}}}}$ are the transformation matrices for textual features of word, cell output of LSTM, representation of attributes learnt from images and videos, respectively, and sigmoid $\sigma$ is element-wise non-linear activation function.

\paragraph{LSTM with Transfer Unit.} Then, we formulate our video captioning with semantic attributes learnt from two sources as a multi-source sequence learning problem and modify the architectures of transfer unit which is treated as the additional input to LSTM for our purpose. The core issue for the modification is about whether the transfer gate in our transfer unit should \emph{individually} or \emph{simultaneously} impact the semantic attributes learnt from different sources. Individual impact means that the transfer gate only critically control the information transferred from attributes in one specific source, while directly leverages the attributes from other source unconditionally. Simultaneous impact decouples the influence of transfer gate such that attributes learnt from different sources can be simultaneously guided with transfer gate.

Our preliminary design LSTM-TSA$_{IV_{0}}$ is the deep fusion without transfer gate by directly utilizing the multimodal layer \cite{Mao:NIPS14}. Specifically, the additional input to LSTM is calculated as
\begin{equation}\label{Eq:Eq0}\small
\mbox{LSTM-TSA$_{IV_{0}}$:}~g\left({{\bf{A_i}}},{{\bf{A_v}}}\right) = {{\bf{T}}_{\bf{A_i}}}{\bf{A_i}}+ {{\bf{T}}_{\bf{A_v}}}{\bf{A_v}},
\end{equation}
where ${{\bf{T}}_{\bf{A_i}}} \in {{\mathbb{R}}^{{D_e} \times {D_{a_i}}}}$ and ${{\bf{T}}_{\bf{A_v}}} \in {{\mathbb{R}}^{{D_e} \times {D_{a_v}}}}$ are the transformation matrices for representation of attributes learnt from images and videos, respectively. Please also note that if only semantic attributes learnt from one single source (images or videos) are available, the additional input $g\left({{\bf{A_i}}},{{\bf{A_v}}}\right)$ to LSTM in our LSTM-TSA will be degraded into $g\left({{\bf{A_i}}}\right)={{\bf{T}}_{\bf{A_i}}}{\bf{A_i}}$ or $g\left({{\bf{A_v}}}\right)={{\bf{T}}_{\bf{A_v}}}{\bf{A_v}}$ and we name these two variants as LSTM-TSA$_{I}$ and LSTM-TSA$_{V}$, respectively.

Then based on the above core design issue, we derive three different architectures of transfer unit as depicted in Figure \ref{fig:figTG}, respectively named as LSTM-TSA$_{IV_{1}}$ to LSTM-TSA$_{IV_{3}}$. The first design (LSTM-TSA$_{IV_{1}}$) individually assigns the attributes learnt from images with the weight vector of transfer gate to dynamically select the favorable information which will be fused as the additional input to LSTM. The second design (LSTM-TSA$_{IV_{2}}$) is similar except that the calculated weight vector of transfer gate is only allocated to the attributes learnt from videos. Both designs are relatively straightforward to implement by multiplying the transformed representation of attributes from one specific source with the weight vector of transfer gate through dot product. The last design (LSTM-TSA$_{IV_{3}}$) is a compromise version between the former two architectures, by simultaneously controlling the two attributes learnt from different sources with decoupled weight vectors from transfer gate, which is also treated as a linear combination between the attributes learnt from images and videos.

Specifically, given the output weight vector ${\bf{g}}^t$ of transfer gate in the time step $t$, the three variants of our transfer unit are designed as
\begin{equation}\label{Eq:Eq1}\small
\mbox{LSTM-TSA$_{IV_{1}}$:}~g\left({{\bf{A_i}}},{{\bf{A_v}}}\right) = {{\bf{T}}_{\bf{A_i}}}{\bf{A_i}}\odot {{\bf{g}}^t}+ {{\bf{T}}_{\bf{A_v}}}{\bf{A_v}},
\end{equation}
\begin{equation}\label{Eq:Eq2}\small
\mbox{LSTM-TSA$_{IV_{2}}$:}~g\left({{\bf{A_i}}},{{\bf{A_v}}}\right) = {{\bf{T}}_{\bf{A_i}}}{\bf{A_i}}+ {{\bf{T}}_{\bf{A_v}}}{\bf{A_v}}\odot {{\bf{g}}^t},
\end{equation}
\begin{equation}\label{Eq:Eq3}\small
\mbox{LSTM-TSA$_{IV_{3}}$:}~g\left({{\bf{A_i}}},{{\bf{A_v}}}\right) = {{\bf{T}}_{\bf{A_i}}}{\bf{A_i}}\odot({1-{\bf{g}}^t})+ {{\bf{T}}_{\bf{A_v}}}{\bf{A_v}}\odot {{\bf{g}}^t},
\end{equation}
where $\odot$ denotes the element-wise dot product function.

\section{Experiments}\label{sec:EX}
We evaluate and compare our proposed LSTM-TSA with state-of-the-art approaches by conducting video captioning task on three video captioning benchmarks, i.e., Microsoft Research Video Description Corpus (MSVD) \cite{Chen:ACL11}, Montreal Video Annotation Dataset (M-VAD) \cite{Torabi15} and MPII Movie Description Corpus (MPII-MD) \cite{Rohrbach:CVPR15}. The first is the most popular video captioning benchmark of YouTube videos and the other two are both recently released large-scale movie description datasets.

\subsection{Datasets}
\textbf{MSVD.} MSVD contains 1,970 video snippets collected from YouTube. There are roughly 40 available English descriptions per video. In our experiments, we follow the setting used in prior works \cite{Guadarrama:ICCV13,Pan:CVPR16}, taking 1,200 videos for training, 100 for validation and 670 for testing.

\textbf{M-VAD.} M-VAD is a recent collection of large-scale movie description dataset. It is composed of about 49,000 DVD movie snippets, which are extracted from 92 DVD movies. Each movie clip is accompanied with single sentence from semi-automatically transcribed descriptive video service (DVS) narrations.

\textbf{MPII-MD.} MPII-MD is another recent collection of movie description dataset, similar to M-VAD. It contains around 68,000 movie snippets from 94 Hollywood movies and each snippet is equipped with a single sentence from movie scripts and DVS.

\subsection{Experimental Settings}
We uniform sample 25 frames/clips for each video and each word in the sentence is represented as ``one-hot" vector (binary index vector in a vocabulary). For video representations, we take the output of 4096-way fc6 layer from the 19-layer VGG \cite{Simonyan14} pre-trained on Imagenet ILSVRC12 dataset \cite{ILSVRC15} and 4096-way fc6 layer from C3D \cite{Tran15} pre-trained on Sports-1M video dataset \cite{KarpathyCVPR14} as frame/clip representation respectively, and concatenate the features from VGG and C3D as the input video representation. For representation of attributes learnt from images, we select the 1,000 most common words on COCO \cite{Lin:ECCV14} as the high-level semantic attributes in the image domain and train the attribute detectors with image MIL model \cite{Fang:CVPR15} purely on the COCO training data, resulting in the final 1,000-way vector of probabilities. For the representation of attributes learnt from videos, 1,000 most common words on each video captioning benchmark are selected individually as semantic attributes in each specific video domain and the corresponding attribute detectors are trained with proposed video MIL model. The dimension of the input and hidden layers in LSTM are both set to 1,024. In testing stage, we adopt the beam search strategy and set the beam size to 4.

For quantitative evaluation of our proposed models, we adopt three common metrics in image/video captioning tasks: BLEU@$N$ \cite{Papineni:ACL02}, METEOR \cite{Banerjee:ACL05}, and CIDEr-D \cite{vedantam2015cider}. All the metrics are computed by using the codes\footnote{\url {https://github.com/tylin/coco-caption}} released by Microsoft COCO Evaluation Server \cite{chen2015microsoft}.

\subsection{Compared Approaches}
To empirically verify the merit of our LSTM-TSA models, we compared the following state-of-the-art methods.

(1) Long Shot-Term Memory (LSTM) \cite{Venugopalan:NAACL15}: LSTM attempts to directly translate from video pixels to natural language with a CNN plus RNN framework. The video representation is generated by performing mean pooling over the frame features across the entire video.

(2) Sequence to Sequence - Video to Text (S2VT) \cite{Venugopalan:ICCV15}: S2VT incorporates both RGB and optical flow inputs, and the encoding and decoding of the inputs and word representations are learnt jointly in a parallel manner.

(3) Temporal Attention (TA) \cite{Yao:ICCV15}: TA combines the frame representation from GoogleNet \cite{Szegedy14} and video clip representation based on a 3-D CNN trained on hand-crafted descriptors. Furthermore, a weighted attention mechanism is exploited to dynamically attend to specific temporal regions of the video while generating sentence.

(4) Long Shot-Term Memory with visual-semantic Embedding (LSTM-E) \cite{Pan:CVPR16}: LSTM-E utilizes both 2-D CNN and 3-D CNN to learn video representation, and simultaneously explores the learning of LSTM and visual-semantic embedding for video captioning.

(5) Convolutional Gated-Recurrent-Unit Recurrent Networks (GRU-RCN) \cite{Ballas:ICLR16}: GRU-RCN leverages convolutional GRU-RNN to extract visual representation and generate sentence based on the LSTM text-generator with soft-attention mechanism \cite{Yao:ICCV15}.

(6) hierarchical Recurrent Neural Networks (h-RNN) \cite{Yu:CVPR16}: Proposed most recently, h-RNN exploits both spatial and temporal attention mechanisms for video captioning.

(7) Hierarchical Recurrent Neural Encoder (HRNE) \cite{pan2015hierarchical}: HRNE encodes the frame sequence with hierarchical RNN and decodes the sentence with attention mechanism.

(8)  Long Short-Term with Transferred Semantic Attributes (LSTM-TSA): We design three runs for our proposed framework, i.e., LSTM-TSA$_{I}$, LSTM-TSA$_{V}$, and LSTM-TSA$_{IV}$. The input semantic attributes of the first two runs LSTM-TSA$_{I}$ and LSTM-TSA$_{V}$ are purely mined from images and videos, respectively. The last run LSTM-TSA$_{IV}$ is to fuse semantic attributes from both images and videos. Note that LSTM-TSA$_{IV_{3}}$ is particularly exploited as LSTM-TSA$_{IV}$ here. The comparisons between four variants of LSTM-TSA$_{IV}$ w or w/o transfer gate will be discussed in Section \ref{ssec:EA}.

\begin{table*}\small
\centering
\caption{METEOR, CIDEr-D, and BLEU@N scores of our LSTM-TSA and other state-of-the-art methods on MSVD dataset. All values are reported as percentage (\%).}
\label{table:MSVD}
\begin{tabular}{l|c|c|c|c|c|c}\hline
~~\textbf{Model}&~~\textbf{METEOR}~~&~~\textbf{CIDEr-D}~~&~~\textbf{BLEU@1}~~&~~\textbf{BLEU@2}~~&~~\textbf{BLEU@3}~~&~~\textbf{BLEU@4}~~
\\ \hline\hline
~~LSTM \cite{Venugopalan:NAACL15} & 29.1 & - &- & - & - & 33.3  \\
~~S2VT \cite{Venugopalan:ICCV15} & 29.8 &- &- & - & - & -  \\
~~TA \cite{Yao:ICCV15}     & 29.6 & 51.7 &80.0 & 64.7 & 52.6 & 41.9  \\
~~LSTM-E \cite{Pan:CVPR16} & 31.0 & - &78.8 & 66.0 & 55.4 & 45.3  \\
~~GRU-RCN \cite{Ballas:ICLR16} & 31.6 & 68.0 &- & - & - & 43.3                \\
~~h-RNN  \cite{Yu:CVPR16} & 32.6 & 65.8 &81.5 & 70.4 & 60.4 & 49.9 \\
~~HRNE \cite{pan2015hierarchical} & 33.1 & - & 79.2 & 66.3 & 55.1 & 43.8 \\\hline
~~\textbf{LSTM-TSA$_{I}$} &32.4 & 71.5 & 81.0 & 69.6 & 60.2 & 50.2 \\
~~\textbf{LSTM-TSA$_{V}$} &32.6 & 71.7 & 82.1 & 70.7 & 61.1 & 50.5 \\
~~\textbf{LSTM-TSA$_{IV}$} &\textbf{33.5} & \textbf{74.0} & \textbf{82.8} & \textbf{72.0} & \textbf{62.8} & \textbf{52.8} \\ \hline
\end{tabular}
\vspace{-0.05in}
\end{table*}

\subsection{Performance Comparison}
\textbf{Quantitative Analysis.} Table \ref{table:MSVD} shows the performances of different models on MSVD dataset. Overall, the results across six evaluation metrics consistently indicate that our proposed LSTM-TSA$_{IV}$ achieves better performance than all the state-of-the-art techniques including non-attention models (LSTM, S2VT, LSTM-E) and attention-based approaches (TA, GRU-RCN, h-RNN, HRNE). In particular, the CIDEr-D of our LSTM-TSA$_{IV}$ can achieve 74.0\% which is to-date the highest performance reported on MSVD dataset, making the relative improvement over TA, GRU-RCN, h-RNN by 43.1\%, 8.8\%, and 12.5\%, respectively. By additionally incorporating attributes to LSTM model, LSTM-TSA$_{I}$ and LSTM-TSA$_{V}$ lead to a performance boost, indicating that visual representations are augmented with high-level semantic attributes and thus do benefit the learning of video sentence generation. As expected, LSTM-TSA$_{V}$ whose attributes are trained in domain outperforms LSTM-TSA$_{I}$ which predicts the attributes learnt on image domain. LSTM-TSA$_{IV}$ utilizing attributes learnt from images and videos significantly improves LSTM-TSA$_{V}$. The result indicates the advantage of leveraging the learnt attributes jointly from two domains which are complementary for boosting video captioning.

\begin{table}
\centering
\caption{\small METEOR (M) scores (\%) of our LSTM-TSA and other state-of-the-art methods on (a) M-VAD and (b) MPII-MD datasets.}
\label{table:Movie}\small
\subtable[M-VAD dataset.]{
\begin{tabular}{@{}l|c@{}}\hline
~~\textbf{Model}~~ & ~~\textbf{M}~~~ \\ \hline\hline
~~TA \cite{Yao:ICCV15}    &  4.3     \\
~~LSTM \cite{Venugopalan:NAACL15}  &  6.1  \\
~~Visual-Labels \cite{Rohrbach:GCPR15} & 6.4\\
~~S2VT \cite{Venugopalan:ICCV15}  &  6.7     \\
~~LSTM-E \cite{Pan:CVPR16}   & 6.7 \\
~~HRNE \cite{pan2015hierarchical}  & 6.8 \\\hline
~~\textbf{LSTM-TSA$_{I}$} & 6.4 \\
~~\textbf{LSTM-TSA$_{V}$} & 6.9 \\
~~\textbf{LSTM-TSA$_{IV}$} & \textbf{7.2} \\\hline
\end{tabular}
}
~~~~
\subtable[MPII-MD dataset.]{
\begin{tabular}{@{}l|c@{}}\hline
~~\textbf{Model}~~ & ~~\textbf{M}~~~ \\ \hline\hline
~~SMT \cite{Rohrbach:CVPR15}  &  5.6      \\
~~LSTM \cite{Venugopalan:NAACL15}  &  6.7 \\
~~Visual-Labels \cite{Rohrbach:GCPR15} & 7.0\\
~~S2VT \cite{Venugopalan:ICCV15}  &  7.1     \\
~~LSTM-E \cite{Pan:CVPR16} & 7.3\\\hline
~~\textbf{LSTM-TSA$_{I}$} & 7.4 \\
~~\textbf{LSTM-TSA$_{V}$} & 7.6 \\
~~\textbf{LSTM-TSA$_{IV}$} & \textbf{8.0} \\\hline
\end{tabular}
}
\vspace{-0.05in}
\end{table}

The performance comparisons in terms of METEOR on two movie datasets M-VAD and MPII-MD are summarized in Table \ref{table:Movie}. The METEOR scores on the two datasets are much lower than those on MSVD, due to the high diversity of visual and textual content in movies. Our LSTM-TSA$_{IV}$ consistently outperforms other baselines in two datasets. The METEOR of LSTM-TSA$_{IV}$ can reach 7.2\% and 8.0\%, which makes the relative improvement over the best competitor HRNE in M-VAD and LSTM-E in MPII-MD by 5.9\% and 9.6\%, respectively. Similar to the observations on MSVD, LSTM-TSA$_{I}$ and LSTM-TSA$_{V}$ exhibit better performance than LSTM by further taking attributes into account for video captioning. In addition, LSTM-TSA$_{V}$ performs better than LSTM-TSA$_{I}$ and larger degree of improvement is attained when exploiting attributes from both images and videos by LSTM-TSA$_{IV}$.

\begin{figure*}[!tb]
\centering {\includegraphics[width=0.97\textwidth]{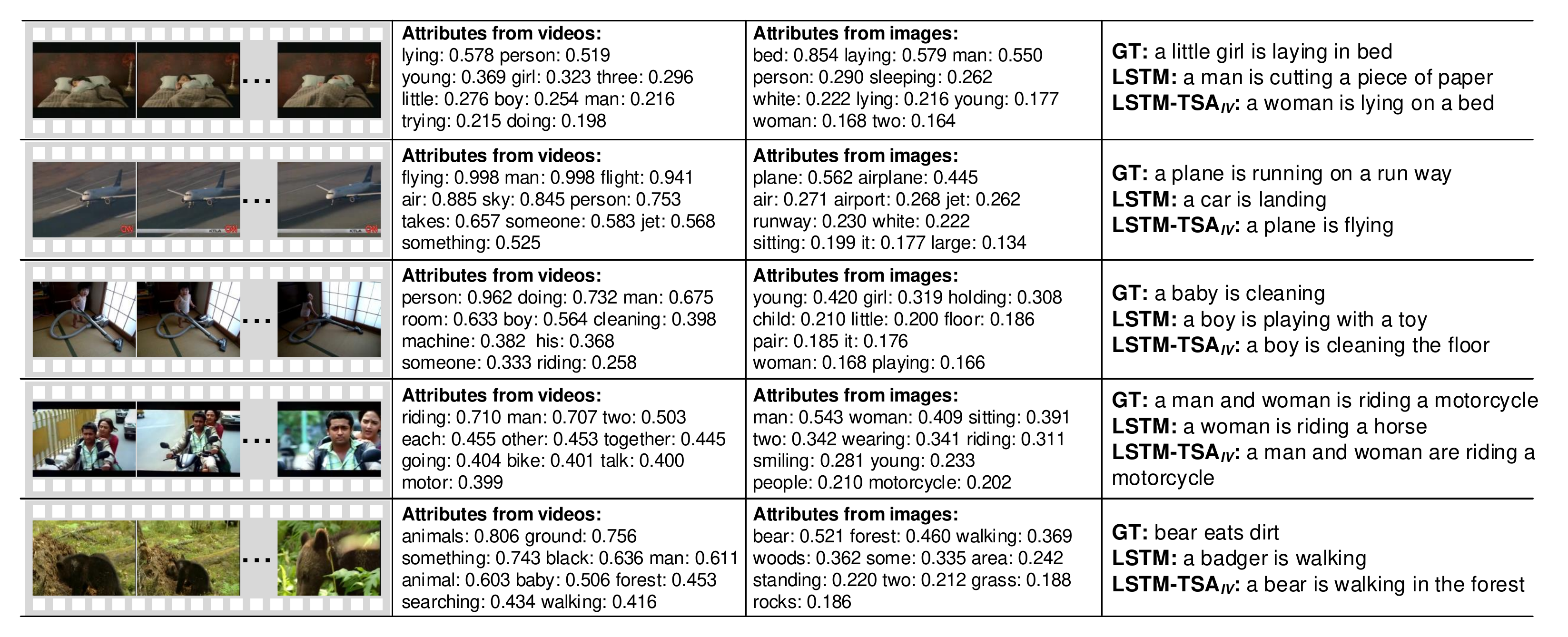}}
\caption{Attributes and sentences generation results on MSVD dataset. The attributes from videos and images are predicted by our video MIL model and image MIL model in \cite{Fang:CVPR15}, respectively, and the output sentences are generated by 1) Ground Truth (GT): One selected ground truth sentence, 2) LSTM, and 3) our LSTM-TSA$_{IV}$.}
\label{fig:figRS}
\vspace{-0.00in}
\end{figure*}

\textbf{Qualitative Analysis.} Figure \ref{fig:figRS} shows a few video examples with the detected semantic attributes from images and videos respectively, human-annotated ground truth sentences and sentences generated by two approaches, i.e., LSTM and our LSTM-TSA$_{IV}$. From these exemplar results, it is easy to see that the two automatic methods can generate somewhat relevant and logically correct sentences, while our model LSTM-TSA$_{IV}$ can predict more accurate words by jointly exploiting video representations and semantic attributes learnt from images and videos for enhancing video captioning. For instance, compared to subject term ``a man" and verb term ``cutting" in the sentence generated by LSTM for the first video, ``a woman" and ``lying" in our LSTM-TSA$_{IV}$ are more relevant to the video content, since the word ``woman" and ``lying" predicted as one attribute from images and videos respectively are directly fed into LSTM to guide the sentence generation. Similarly, verb term ``cleaning" detected as an attribute from videos and object term ``floor" learnt from images present the third image more exactly. Moreover, our LSTM-TSA$_{IV}$ can generate more descriptive sentence by enriching the semantics with attributes. For instance, with the detected term ``forest," the generated sentence ``a bear is walking in the forest" of the fifth video depicts the video content more comprehensive. This confirms that video captioning is benefited by leveraging complementary attributes learnt from images and videos.

\subsection{Experimental Analysis}\label{ssec:EA}
We further verify the effectiveness of our proposed video MIL framework for attribute learning and compare the different variants of our designed transfer unit.

\textbf{Evaluation of Video MIL Framework.} There are generally two directions for attribute learning on videos. One is to perform image MIL model on individual video frame and the other is our proposed video MIL model to jointly utilize all the sampled frames from one video, as shown in Figure \ref{fig:figMIL}. Table \ref{table:FV} compares the sentence generation performances of the LSTM-TSA$_{V}$ model with semantic attributes only learnt from videos by these two different MIL models on MSVD dataset. The results across different metrics consistently indicate that LSTM-TSA$_{V}$ with semantic attributes learnt by video MIL model leads to a better performance, demonstrating the advantage of exploring semantic information among all the sampled frames from one video holistically, as opposed to locally based on individual frame.

\begin{table}\small
\centering
\caption{\small METEOR, CIDEr-D, and BLEU@4 scores of our proposed model LSTM-TSA$_{V}$ with semantic attributes only learnt from videos by two different MIL models on MSVD dataset. One is to perform image MIL model on individual video frame and the other is our proposed video MIL model as shown in Figure \ref{fig:figMIL}. All values are reported as percentage (\%).}
\label{table:FV}
\begin{tabular}{l|c|c|c}\hline
~~\textbf{Model}&~\textbf{METEOR}~&~\textbf{CIDEr-D}~&~\textbf{BLEU@4}~\\ \hline\hline
~~\textbf{Image MIL model} &32.0 & 70.6 & 48.8 \\
~~\textbf{Video MIL model} &\textbf{32.6} & \textbf{71.7} & \textbf{50.5} \\\hline
\end{tabular}
\vspace{-0.05in}
\end{table}

\textbf{Evaluation of Transfer Unit.} Next, we turn to evaluate different variants of our designed transfer unit towards sentence generation. The performances on MSVD dataset of our LSTM-TSA$_{IV}$ are shown in Table \ref{table:TG}, by combining attributes learnt from images and videos with different variants of transfer unit. LSTM-TSA$_{IV_{0}}$ directly calculates an element-wise sum of the feature mappings of attributes from images and videos as a combination, which is fed into LSTM as an additional input. Thus, this additional input is shared and fixed at each time step in LSTM. In contrast, LSTM-TSA$_{IV_{1}}$, LSTM-TSA$_{IV_{2}}$ and LSTM-TSA$_{IV_{3}}$ fuses the two attributes with a transfer gate that dynamically computes a distinct weight based on the two attributes, the current input word and the previous hidden state in LSTM, and then computes the additional inputs to LSTM by applying the weight to attributes from images, videos and both, respectively. As such, the weight offers a more precise control of impacts from semantic attributes by integrating context information and is different at each time step. As indicated by our results, utilizing transfer gate which dynamically balances the influence between attributes learnt from images and videos can constantly lead to better performance than LSTM-TSA$_{IV_{0}}$. A larger performance gain is attained when applying the weight on attributes from both.

\begin{table}\small
\centering
\caption{METEOR, CIDEr-D, and BLEU@4 scores of our proposed model LSTM-TSA$_{IV}$ with semantic attributes learnt from both images and videos on MSVD dataset. Results are shown utilizing the different input architectures of LSTM w/o transfer gate.}
\label{table:TG}
\begin{tabular}{l|c|c|c}\hline
~~\textbf{Model}&~~\textbf{METEOR}~~&~~\textbf{CIDEr-D}~~&~~\textbf{BLEU@4}~~\\ \hline\hline
~~\textbf{LSTM-TSA$_{IV_{0}}$} &32.7 & 71.7 & 50.3 \\
~~\textbf{LSTM-TSA$_{IV_{1}}$} &32.9 & 71.5 & 51.2 \\
~~\textbf{LSTM-TSA$_{IV_{2}}$} &33.0 & 72.3 & 50.5 \\
~~\textbf{LSTM-TSA$_{IV_{3}}$} &\textbf{33.5} & \textbf{74.0} & \textbf{52.8} \\\hline
\end{tabular}
\vspace{-0.14in}
\end{table}

\section{Discussions and Conclusions}\label{sec:CON}
We have presented Long Short-Term Memory with Transferred Semantic Attributes (LSTM-TSA) architecture which explores both video representations and semantic attributes for video captioning. Particularly, we study the problems of how to mine attributes from images and videos and how to fuse them in an elegant manner for enhancing sentence generation. To verify our claim, we have presented video MIL framework to holistically explore semantic information in a video and a transfer unit to contextually control the impacts of attributes learnt from images and videos. Experiments conducted on three widely adopted video captioning datasets validate our proposal and analysis. Performance improvements are clearly observed when comparing to other captioning techniques.

Our future works are as follows. First, attention mechanism will further be incorporated into our LSTM-TSA architecture for further boosting video captioning. Second, we will investigate how to leverage semantic attributes for multiple sentence or paragraph generation for videos.

{\small
\bibliographystyle{ieee}
\bibliography{egbib}
}

\end{document}